\documentclass[sigconf]{acmart}
\usepackage{multicol}
\usepackage{multirow}
\usepackage{subcaption}
\usepackage{balance} 
\AtBeginDocument{%
  }

\setcopyright{acmlicensed}
\copyrightyear{2025}
\acmYear{2025}
\acmDOI{https://doi.org/10.1145/3746027.3755635}
\acmConference[MM '25] {Proceedings of the 33rd ACM International Conference on Multimedia}{October 27--31, 2025}{Dublin, Ireland}
\acmBooktitle{Proceedings of the 33rd ACM International Conference on Multimedia (MM '25), October 27--31, 2025, Dublin, Ireland}
\acmISBN{979-8-4007-2035-2/2025/10}

\begin{document}

\title{AF-CLIP: Zero-Shot Anomaly Detection via Anomaly-Focused CLIP Adaptation}


\author{Qingqing Fang}
\orcid{0009-0006-9876-9863}
\affiliation{%
  \department{School of Computer Science and Engineering}
  \institution{Sun Yat-sen University}
  \city{Guangzhou}
  \country{China}
}
\email{fangqq3@mail2.sysu.edu.cn}

\author{Wenxi Lv}
\affiliation{%
  \department{School of Computer Science and Engineering}
  \institution{Sun Yat-sen University}
  \city{Guangzhou}
  \country{China}
}
\email{lvwx8@mail2.sysu.edu.cn}

\author{Qinliang Su}
\authornote{Corresponding author.}
\affiliation{%
  \department{School of Computer Science and Engineering}
  \institution{Sun Yat-sen University}
  \city{Guangzhou}
  \country{China}
}
\affiliation{%
  \institution{Guangdong Key Laboratory of Big Data Analysis and Processing}
  \city{Guangzhou}
  \country{China}
}
\email{suqliang@mail.sysu.edu.cn}

\renewcommand{\shortauthors}{Trovato et al.}

\begin{abstract}
  Visual anomaly detection has been widely used in industrial inspection and medical diagnosis. Existing methods typically demand substantial training samples, limiting their utility in zero-/few-shot scenarios. While recent efforts have leveraged CLIP's zero-shot recognition capability for this task, they often ignore optimizing visual features to focus on local anomalies, reducing their efficacy. In this work, we propose AF-CLIP (Anomaly-Focused CLIP) by dramatically enhancing its visual representations to focus on local defects. Our approach introduces a lightweight adapter that emphasizes anomaly-relevant patterns in visual features, simultaneously optimizing both class-level features for image classification and patch-level features for precise localization. To capture anomalies of different sizes and improve detection accuracy, prior to the adapter, we develop a multi-scale spatial aggregation mechanism to effectively consolidate neighborhood context. Complementing these visual enhancements, we design learnable textual prompts that generically characterize normal and abnormal states.  After optimization on auxiliary datasets using a composite objective function, AF-CLIP demonstrates strong zero-shot detection capability. Our method is also extended to few-shot scenarios by extra memory banks. Experimental results across diverse industrial and medical datasets demonstrate the effectiveness and generalization of our proposed method.
  Code is available at \url{https://github.com/Faustinaqq/AF-CLIP}.
\end{abstract}

\ccsdesc[500]{Computing methodologies~Scene anomaly detection}

\keywords{Visual Anomaly Detection, Zero-shot, Few-shot, CLIP}



\maketitle

\section{Introduction}

\begin{figure}[tbp]
  \centering
  \begin{subfigure}[b]{0.058\textwidth}
    \includegraphics[width=\textwidth]{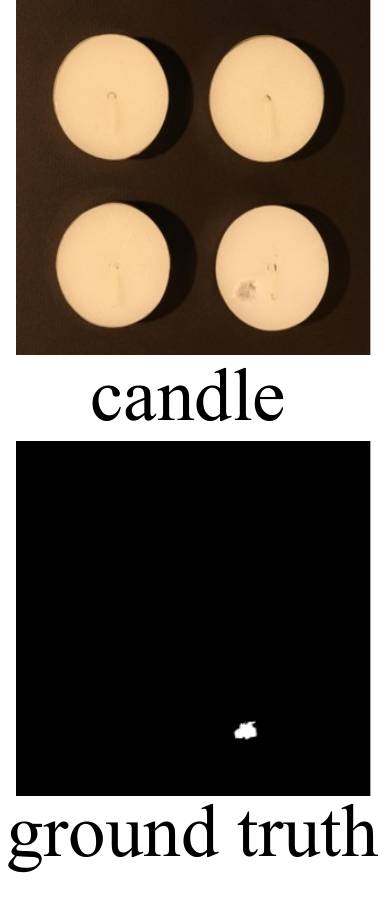}
    \caption{Img}
    \label{fig:attn_image}
  \end{subfigure}
  \hfill
  \begin{subfigure}[b]{0.132\textwidth}
    \includegraphics[width=\textwidth]{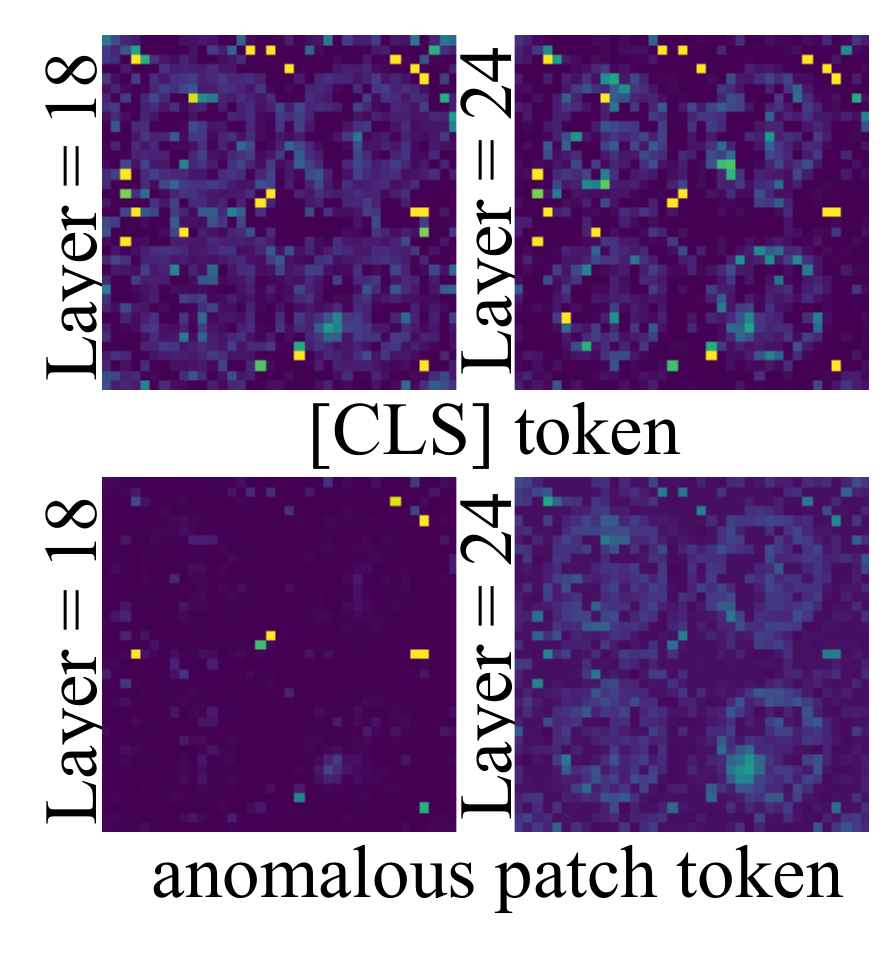}
    \caption{Original CLIP}
    \label{fig:ori_attn}
  \end{subfigure}
  \hfill
  \begin{subfigure}[b]{0.132\textwidth}
    \includegraphics[width=\textwidth]{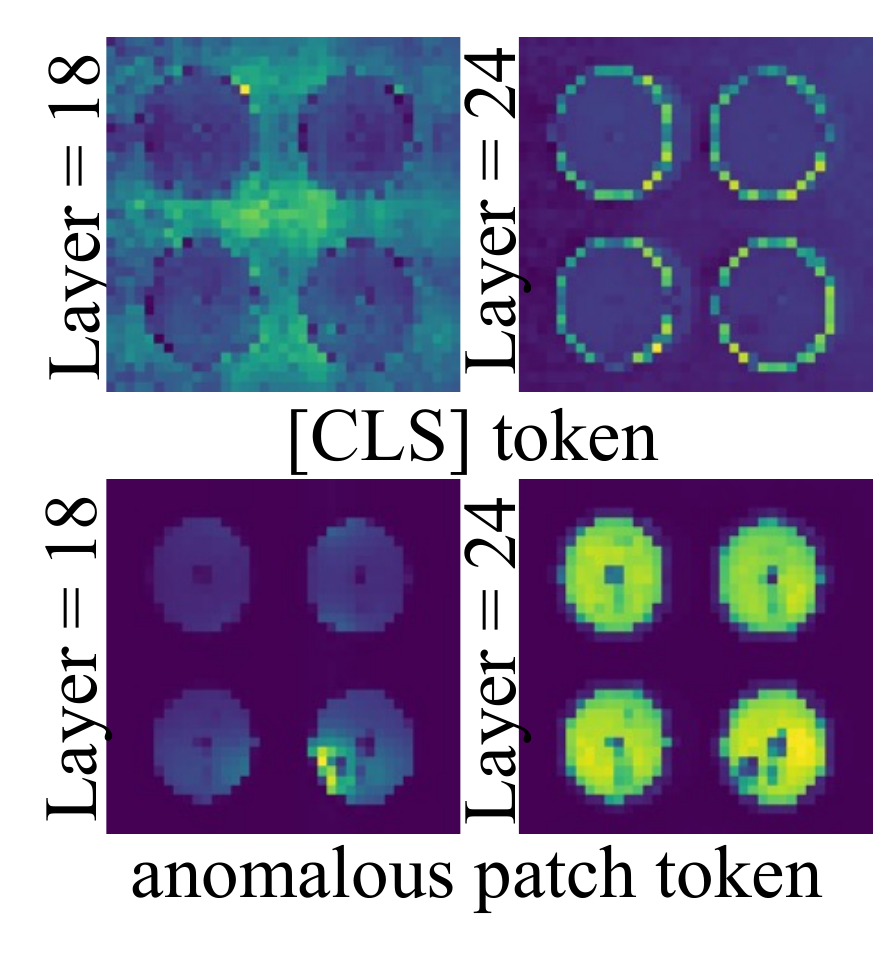}
    \caption{VV-attention}
    \label{fig:vv_attn}
  \end{subfigure}
  \hfill
  \begin{subfigure}[b]{0.132\textwidth}
    \includegraphics[width=\textwidth]{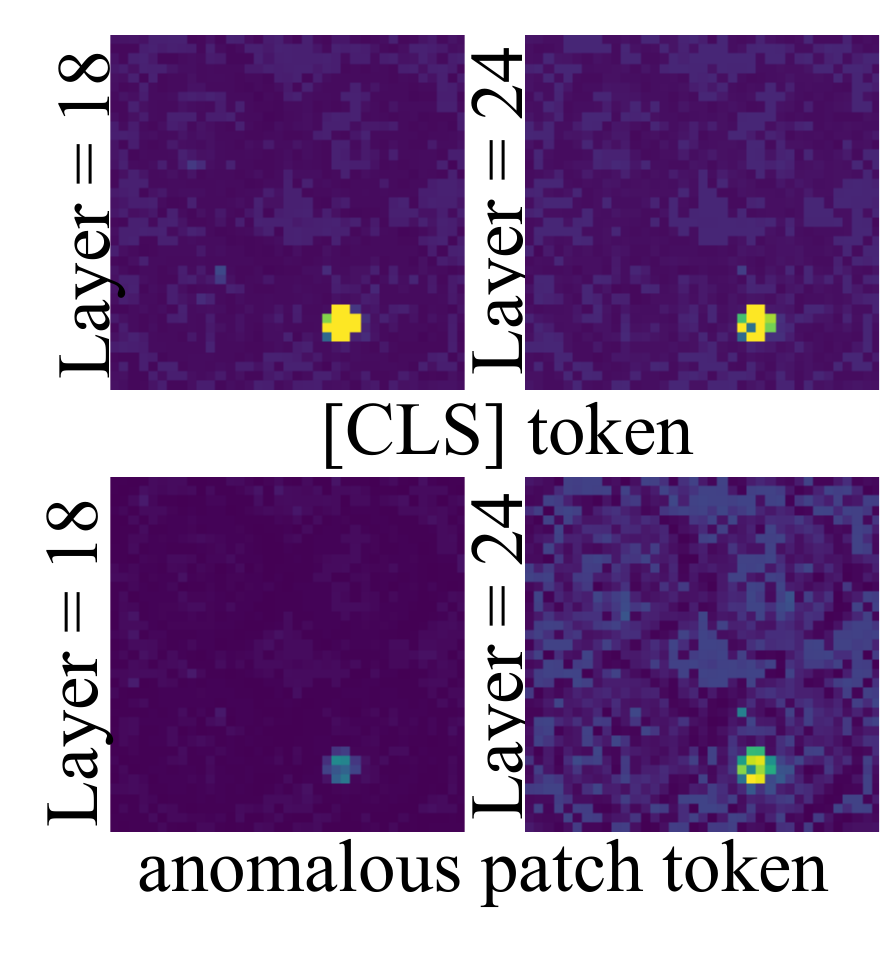}
    \caption{AF-CLIP}
    \label{fig:friclip_attn}
  \end{subfigure}
  \caption{Illustration of attention maps of [CLS] token and anomalous patch token in the CLIP visual encoder. In the original CLIP, [CLS] token broadly attends to the entire image rather than the defective area, and same is observed for the anomalous patch token. The VV-attention helps patch tokens capture local anomalies in shallow layers but loses this capability in deeper layers, while the [CLS] token is always unable to attend to the defective areas. In contrast, in AF-CLIP, both the class token and the anomalous patch token are effectively redirected to focus on the anomalous area.}
  \label{fig:attn}
  \Description{attention map in origin CLIP and AF-CLIP.}
\end{figure}

Visual anomaly detection, which aims to identify anomalous samples and locate their areas within images, consists of two subtasks: anomaly classification and anomaly segmentation. It has been widely used in various applications, such as industrial inspection~\cite{deng2022RD,roth2022patchcore,tien2023revisiting} and medical diagnosis~\cite{zhao2023aeflow,zhou2024anomalyclip}. 
Due to the diversity of anomalies and the inherent difficulties of collecting and annotating anomalies, most existing visual anomaly detection methods~\cite{zavrtanik2021reconstruction, deng2022RD, tien2023revisiting,roth2022patchcore, reiss2023mean, hyun2024reconpatch,zavrtanik2021draem, liu2023simplenet} follow the unsupervised paradigm,  assuming access to a substantial number of normal samples from the target domain.
However, this assumption may not always hold in real world since collecting extensive training samples could be challenging for various reasons. For example, we may not be able to access some patients' data due to privacy protection; and when producing an entirely new product in industrial production, there is no training data available. Under these scenarios, there is an urgent need for the zero-shot anomaly detection paradigm, which does not require to use any training data from the target domain. 


To achieve zero-shot anomaly detection, existing works primarily make use of vision-language models like CLIP ~\cite{clip,li2022blip,li2023blip,kirillov2023segment} owing to their strong zero-shot generalization ability ~\cite{jeong2023winclip,chen2023VAND,zhou2024anomalyclip}. For instance, WinCLIP~\cite{jeong2023winclip} detects anomalies by assessing the alignment between visual features of image patches and the embedding of hand-crafted textual prompts that describe possible anomaly states. When an alignment between an image patch and an anomaly prompt is confirmed, the patch is considered to be flawed. However, due to the difficulties of crafting textual prompts that cover all kinds of possible anomaly states, it brings significant inconveniences for real-world applications.  To overcome the problem, motivated by the success of prompt learning~\cite{lester2021prompt}, many recent works propose to directly learn textual prompts from data rather than manually crafting them, such as  AnomalyCLIP~\cite{zhou2024anomalyclip}, AdaCLIP~\cite{cao2024adaclip}, PromptAD~\cite{li2024promptad} and {\it etc.}. Under the zero-shot paradigm, these works generally use anomaly datasets collected from other domains to help learn the prompts, which are then transferred to the target domain directly. The rationale behind this transfer learning approach is that anomalies (e.g., scratches, dents, and structural distortions) exhibit consistent characteristics across different objects, enabling the learned prompts to be effectively utilized across domains.

Existing works primarily focus on the designing and learning of textual prompts, while leaving how to obtain effective visual features largely unexplored. However, since CLIP is trained on large-scale natural image-text pairs through contrastive learning, its visual encoder is primarily trained for recognition of main objects in images, rather than the local subtle anomalies. Consequently,  [CLS] and patch tokens distribute their focus mainly on the global structures, rather than the more important local defective areas for anomaly detection tasks, as shown in Figure \ref{fig:attn}(b). To alleviate the problem, AnomalyCLIP~\cite{zhou2024anomalyclip} and PromptAD~\cite{li2024promptad} propose to replace the original Q-K-V attention in CLIP with the V-V attention~\cite{li2023clip-surgery} during the forward computation of visual features. Despite the V-V attention can partially increase the model's attention to local areas, it also limits the model's ability to attend to distant regions. This could be problematic for learning the feature of [CLS] token and detecting non-local defects like scratches since both require to gather information from distant regions. Moreover, we also find that V-V attention may attend to very broad areas in deep layers, reducing its effectiveness for anomaly detection, as shown in Figure \ref{fig:attn}(c). In addition to the usage of V-V attention, some works like VAND\cite{chen2023VAND}, AdaCLIP~\cite{cao2024adaclip} and AA-CLIP~\cite{ma2025aaclipenhancingzeroshotanomaly} also propose to add a trainable projection layer onto each visual token separately to refine the visual features so that they can align with the textual prompts more easily. However, this does not allow visual features of the [CLS] or anomalous token to gather information from distant anomalous patches. For anomaly detection tasks, it is important to ensure visual features to mainly contain anomalous information, rather than the object information as in classification tasks.

In this paper, we propose to introduce a lightweight attention adapter at the output of CLIP's layers to increase the vision encoder's attention on anomalous areas. By finetuning the attention adapter on an auxiliary anomaly dataset as in previous works ~\cite{zhou2024anomalyclip,cao2024adaclip,ma2025aaclipenhancingzeroshotanomaly}, the [CLS] token is explicitly guided to attend to the distant anomalous areas, while the anomalous patch tokens are taught to focus more on local anomalous regions, as evidenced in Figure \ref{fig:attn}(d). In this way, we obtain anomaly-rich  [CLS] and patch features, making them suitable for anomaly detection and localization. To account for different scales of anomalies as well as explicitly impose spatial information into the features, spatial aggregation is performed before feature adaptation by multi-scale sliding windows, aiming to capture crucial contextual information for precise anomaly localization. Additionally, a patch alignment loss is further used to ensure the distinction between the visual features of normal and abnormal patches.  Besides these visual adaptations, prompt learning technique is used to capture the generic normal and abnormal textual states learned from the auxiliary dataset. After training the adapter and prompts, anomaly scores are obtained by evaluating the similarity between textual and visual features for zero-shot anomaly detection. Later, we further extend the model to the few-shot setting, under which several normal images are assumed to be available, and introduce the AF-CLIP+ by using a memory bank to store the visual features of normal patches. Experiments conducted on multiple industrial and medical datasets show the effectiveness and superiority of our proposed method. Our method achieves zero-shot image-level/pixel-level AUROC 92.9\%/92.3\% on MVTec, 88.5\%/96.2\% on Visa, and 97.2\%/97.6\% on MVTec, 92.2\%/98.5\% on Visa for 4-shot anomaly detection, surpassing some recent state-of-the-art works.

To summarize, the key contributions of this paper include:
\begin{itemize}
    \item We propose AF-CLIP, a zero-shot anomaly detection framework that adapts CLIP to focus on local anomalies by a lightweight adapter and spatial aggregation, and its normal/abnormal visual features are forced to be discriminative by using a patch alignment loss.
    \item We extend AF-CLIP to the few-shot method AF-CLIP+, which introduces memory banks to store aggregated visual features from normal samples, enabling robust anomaly detection under few-shot scenarios.
    \item Extensive experiments on multiple industrial and medical datasets show the effectiveness of our zero-shot model, while the few-shot performance on MVTec and Visa benchmarks also surpasses some recent state-of-the-art methods.
\end{itemize}

\section{Releated Work}

\paragraph{Vision-Language Model.} Based on transformer ~\cite{vaswani2017attention} and vision transformer (ViT)~\cite{dosovitskiy2020vit} architecture, recent visual-language models (VLM) ~\cite{clip,li2022blip,li2023blip} trained on large-scale image-text pairs collected from the website, exhibit remarkable generality in zero-shot text-image inference. Among them, CLIP~\cite{clip} is a widely used VLM, which learns a joint vision-language feature space via contrastive learning through its text and image encoders. Benefiting from its strong zero-shot classification ability and successful fine-tuning methods like prompt tuning~\cite{lester2021prompt,liu2022promptv2},prefix-tuning~\cite{li2021prefix}, and LoRA~\cite{hu2022lora}, CLIP can be adapted well for downstream tasks by just fine-tuning a small number of parameters. 

\paragraph{Anomaly Detection}
Due to the limited anomalies, most anomaly detection methods follow the unsupervised paradigm, training with only normal samples. Among them, reconstruction-based methods~\cite{ZhangWC22,zavrtanik2021reconstruction, deng2022RD,tien2023revisiting,fang2025boosting} reconstruct normal images or features extracted from pre-trained networks, assuming that anomalous samples can not be reconstructed well, and detect anomalies by the reconstructed error. Embedding-based methods ~\cite{yi2020p-svdd,gudovskiy2022cflow,reiss2023mean,roth2022patchcore,hyun2024reconpatch} focus on learning compact feature spaces for normal samples, identifying anomalies via distinctive feature deviations.
Pseudo-based methods~\cite{zavrtanik2021draem,li2021cutpaste,liu2023simplenet,zhang2023destseg} synthesize pseudo-anomalies and masks from normal samples to directly train segmentation networks. While effective in detecting anomalies, these unsupervised methods typically need to train a separate model for each category based on lots of normal samples, costing much memory and time.


\paragraph{Zero-/Few-shot Anomaly Detection} With only zero-/few-shot samples, traditional anomaly detection methods, which rely on large amounts of samples, can not work well. To classify and localize anomalies under such scenarios, recent works leverage CLIP~\cite{clip} for the downstream task. WinCLIP \cite{jeong2023winclip} uses manually designed text prompts to describe normality/anomality for each category, splits images into multi-scale patches for visual features, and matches text-visual features for anomaly identification. VNAD \cite{chen2023VAND} employs four linear projectors to map image features to text space, achieving strong performance. Besides, inspired by CoOp \cite{zhou2022learningcoop} and CLIP-surgery \cite{li2023clip-surgery},
AnomalyCLIP \cite{zhou2024anomalyclip} introduces learnable object-agnostic text prompts and diagonally prominent attention to enhance zero-shot capability. AdaCLIP~\cite{cao2024adaclip} adapts dual hybrid learnable prompts, SimCLIP \cite{deng2024simclip} fine-tunes both vision and text embeddings of CLIP, and AA-CLIP~\cite{ma2025aaclipenhancingzeroshotanomaly} uses two-stage training strategies to transfer CLIP for anomaly detection. For few-shot scenarios, PromptAD \cite{li2024promptad} learns semantic concatenation of category-specific normal/abnormal prompts; while IIPAD \cite{lvone} proposes instance-induced prompt learning for a unified "one-for-all" model.  These methods have shown the effectiveness and applicability of vision-language models in anomaly detection, but their simple adaptation of visual features still remains improvement space for later works.

\section{Preliminaries of CLIP}
Contrastive Language-Image Pre-training (CLIP)~\cite{clip} is well-known for its ability to produce text-image aligned representations via its textual and image encoders $f_T(\cdot)$ and $f_I(\cdot)$. Capitalizing on this alignment ability, given an image $x$ and different texts $\{t_k\}_{k=1}^K$ which describe the categories like ``\texttt{a photo of a [c].}'' where ``\texttt{[c]}'' is the class name, the probability of an image belong to class $c_k$ can be obtained by 
\begin{equation}
     p(x \in c_k) = \frac{exp(\langle f_{I}(x), f_{T}(t_k)\rangle / \tau)}{\sum_{k=1}^K exp(\langle f_{I}(x), f_{T}(t_k)\rangle/\tau)},
\end{equation}
where $\langle \cdot, \cdot\rangle$ is the cosine similarity of two vectors, and $\tau$ is the temperature hyperparameter. Despite its success in general vision-language tasks, CLIP exhibits critical limitations when applied to anomaly detection due to its pre-trained visual features being optimized for holistic appearance recognition rather than local anomaly identification. 
To address this fundamental mismatch, we present AF-CLIP to bridge this gap by transforming CLIP's visual representations into anomaly-focused features.

\section{Methodology}

\begin{figure*}[htbp]
    \centering
    \includegraphics[width=0.85\linewidth]{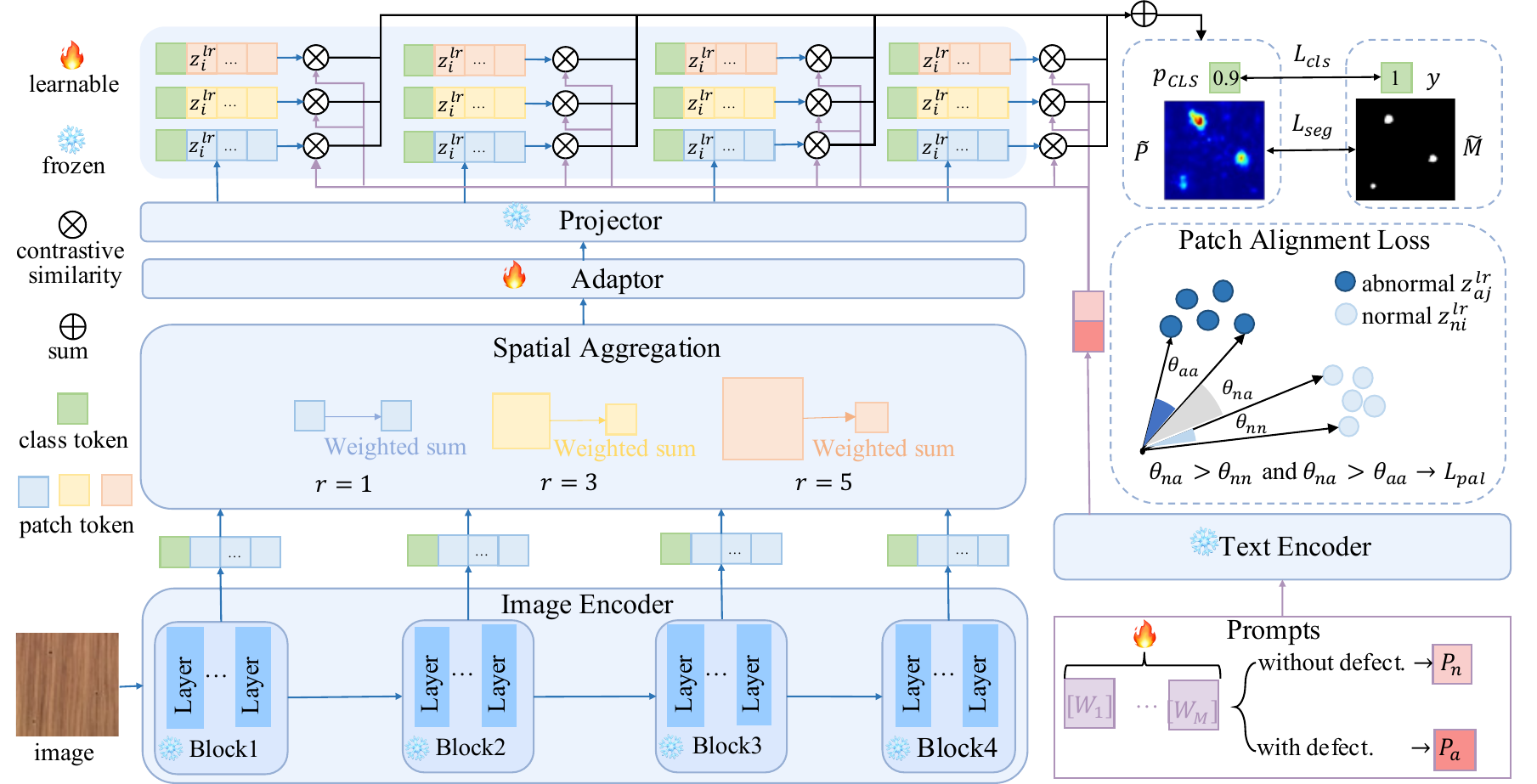}
    \caption{Structure of AF-CLIP. Visual Features from different blocks are aggregated with different scale spatial information and then refined to be anomaly-focused by the adaptor. These visual features are then aligned with learnable textual prompts and optimized on the auxiliary dataset by $L_{cls},L_{seg}$ and patch alignment loss $L_{pal}$ for zero-shot anomaly detection.}
    \label{fig:structure}
    \Description{Structure of proposed AF-CLIP} 
\end{figure*}

As shown in Figure \ref{fig:structure}, AF-CLIP is proposed to adapt CLIP for zero-shot visual anomaly detection. We mainly adjust visual features by spatial information aggregation and an additional adapter, then align them with generic textual prompts. After optimizing AF-CLIP on auxiliary anomaly datasets by classification loss, segmentation loss and patch alignment loss, it can be used to detect anomalies under zero-shot scenarios.
In the following, we will introduce each component one by one.

\subsection{Anomaly-Focused Visual Feature Adaption}


CLIP's image encoder produces visual features targeting global semantic recognition of objects rather than localized anomaly detection. To bridge this mismatch, we propose to introduce an adapter as well as a spatial aggregator to explicitly redirect visual features to focus on anomaly-relevant patterns.

\paragraph{Anomaly-Aware Feature Adaptation}
As done in previous works ~\cite{chen2023VAND,zhou2024anomalyclip,li2024promptad}, we partition the ViT-based image encoder into four hierarchical blocks $\mathcal{L}=\{1,2,3,4\}$ and extract visual features from each block $\ell \in \mathcal{L}$ for an input image $x$:
\begin{align}
    a^{\ell} = [a^\ell_{CLS}, a^\ell_1, \dots, a^\ell_N] = f_I^{\ell}(x) \in \mathbb{R}^{(N+1) \times d},
\end{align}
where $a^\ell_{CLS}$ denotes the global class token and $\{a^\ell_i\}_{i=1}^N$ represent local patch features.
To explicitly redirect the model's attention to potential anomaly regions, we introduce a lightweight yet effective adaptor module based on trainable attention mechanisms~\cite{vaswani2017attention}. The adaptor operates through three key transformations:
\begin{align}
    Q^{\ell} = a^{\ell} \cdot W_{Q}, \quad K^{\ell} = a^{\ell} \cdot W_{K}, \quad V^{\ell} = a^{\ell} \cdot W_{V},
\end{align}
where $W_{Q}, W_{K}, W_{V} \in \mathbb{R}^{d \times \tilde{d}}$ are learnable projection matrices. The core innovation lies in the subsequent attention operation:
\begin{align}
    \tilde{a}^{\ell} = \text{Attention}(Q^{\ell}, K^{\ell}, V^{\ell}) \cdot W_O.
\end{align}
By training on an auxiliary anomaly dataset, the attention-based adaptor can achieve two crucial goals:
\begin{enumerate}
    \item The class token $a^\ell_{CLS}$ is encouraged to attend to anomaly-relevant patches, effectively aggregating local defect patterns for anomaly detection, as the first row of Figure\ref{fig:attn}(d) shows;
    \item Patch features $\{a^\ell_i\}$ is redirected to focus on local anomalous information, enabling precise anomaly localization for segmentation, as the second row of Figure\ref{fig:attn}(d) shows.
\end{enumerate}
By having the original CLIP features from different layers go through the simple adaptor, we can obtain visual features that are more suitable for both anomaly detection and segmentation tasks. For simplicity, we denote the adjustment of the adaptor as $\tilde{a}^{\ell} = {\mathcal{H}}_a(a^\ell)$.
\paragraph{Spatial Information Aggregation}

While the adaptor effectively extracts anomaly-relevant features, the fixed patch size (e.g., 14×14 or 16×16 pixels) in ViT-based encoders limits their ability to detect defects of different scales. Moreover, unlike CNNs which inherently take local spatial relationships into account, Vision Transformers lack strong inductive bias for modeling neighborhood context, which is crucial for precise anomaly localization and classification.
To address these limitations, we introduce a multi-scale spatial aggregation module before the adaptor. The process begins by reshaping the patch features $a^\ell$ into spatial feature maps $b^\ell \in \mathbb{R}^{H\times W\times d}$ (where $H = W = \sqrt{N}$). We then perform the spatial aggregation under different window sizes $r \in \mathcal{R} = \{1,3,5\}$, respectively. Under the $r\times r$ windows, to obtain neighbor information as well as retain the center patch information, we propose to perform a weighted average over features within a $r\times r$ window. Specifically, for the patch feature $b^{\ell}_{hw}$, the weight of its neighbor $b^{\ell}_{ij}$ is designated as
$
    \eta_{ij}^{r} = \frac{\eta_{ij}}{\sum_{i'=h-\lfloor \frac{r}{2} \rfloor}^{h+\lfloor \frac{r}{2} \rfloor}\sum_{j'=w-\lfloor \frac{r}{2} \rfloor}^{w+\lfloor \frac{r}{2} \rfloor}\eta_{i'j'}}
$,
where $\eta_{ij} = \exp(-\frac{(i-h)^2 + (j-w)^2}{2\sigma^2})$, and $\sigma$ is the standard deviation.
The aggregated feature at each position is then calculated through the weighted summation:
\begin{align}
    \tilde{b}^{\ell r}_{h w} = \sum_{i=h-\lfloor \frac{r}{2} \rfloor}^{h+\lfloor \frac{r}{2} \rfloor}\sum_{j=w-\lfloor \frac{r}{2} \rfloor}^{w+\lfloor \frac{r}{2} \rfloor}\eta^{r}_{ij} b^{\ell}_{ij}
\end{align}
The aggregated features are reformulated as $\tilde{b}^{\ell r} \in \mathbb{R}^{N \times d}$ and combined with the class token, forming multi-scale visual features:
\begin{align}
   c^{\ell r} = [a^\ell_{CLS};\tilde{b}^{\ell r}]; \quad \mathcal{C} = \left\{c^{\ell r} | l \in \mathcal{L} , r\in \mathcal{R} \right\}.
   \label{eq:agg}
\end{align}
Notably, when $r=1$, $c^{\ell r}$ is actually the original feature $a^\ell$. But for $r\ne1$, $c^{\ell r}$ contains the neighborhood information of each patch. Thus, in our final model, we feed the spatial-ware features $c^{\ell r}$ for $r \in \{1, 3, 5\}$ into the adator ${\mathcal{H}}_a(\cdot)$ to extract anomaly-relevant visual features as $\tilde{c}^{\ell r} = {\mathcal{H}}_a(c^{\ell r})$. In this way, the anomaly-focused information is further emphasized at different scales.

Finally, the visual features $\tilde{c}^{\ell r}$ are further projected into the joint text-image space via the projection head $f_p(\cdot)$ in original CLIP:
\begin{align}
z^{\ell r}=f_p(\tilde{c}^{\ell r}) \in \mathbb{R}^{(N+1) \times t}, \quad \mathcal{Z}=\left\{z^{\ell r} | \ell \in \mathcal{L}, r \in \mathcal{R}\right\},
\end{align}
where $t$ is the dimension of the final visual features.

\subsection{Vision-aligned Textual State Prompts}
Besides the visual features, textual features also play a vital role in identifying anomalies.
Instead of manually designing category-specific prompts, we propose learning generic text representations. 
Our approach leverages auxiliary anomaly datasets containing objects with possible shared defect patterns (e.g., scratches, holes).
To focus the learning on generic normal/abnormal state representations rather than dataset-specific characteristics, we design prompts with fixed state-descriptive suffixes and shared learnable parameters:
\begin{align}
    [W_1][W_2][\dots][W_{M}] + \left\{
\begin{aligned}
& \text{without defect.} \rightarrow P_n, \\
& \text{with defect.} \rightarrow P_a,
\end{aligned}
\right.
\end{align}
where $[W_i]$ are learnable word embeddings and $M$ is the length of learnable tokens. These shared, learnable prefixes adapt to the fixed state descriptors (``without/with defect''), forcing the model to focus on the fundamental normal-abnormal semantic distinction rather than dataset-specific features. 
We pass these prompts through the text encoder to get the normal and anomalous representations, denoted as 
$t_n = f_T(P_n), t_a = f_T(P_a)$, 
which are then aligned with visual features to output anomaly scores.

\subsection{Training of Anomaly-Focused CLIP}
To drive the visual features in AF-CLIP to focus on local anomalies, as done in previous works, we assume the availability of an auxiliary anomaly detection dataset $\{\mathcal{X}, \mathcal{Y}, \mathcal{M}\}$. The dataset includes normal and anomalous images $x \in \mathcal{X}$, as well as their image-level labels $y \in \mathcal{Y}$ and pixel-level labels $M \in \mathcal{M}$. With frozen pre-trained CLIP, we train the parameters from the adaptor and learnable prompt embeddings. After training, the visual features are encouraged to be aware of local anomalous patches, thus can be used to compute the alignment with textual features. 
Specifically, for any visual feature $z^{\ell r}_i$, the abnormal and normal similarity can be obtained by
\begin{equation}
    S^{\ell r}_{ia}\!=\!\frac{\langle z^{\ell r}_i\!,t_a \rangle}{\tau},\quad
    S^{\ell r}_{in}\!=\!\frac{\langle z^{\ell r}_i\!, t_n \rangle}{\tau},
    \quad i \in \{CLS, 1, \dots, N\},
\end{equation}
where $\langle\cdot\rangle$ is the cosine similarity. Summing the score over all blocks $\ell$ and scales $r$ for each token $i$, the whole score can be denoted as $S_{ia} = \sum_{\ell, r} S^{\ell, r}_{ia}$, $S_{in} = \sum_{\ell r} S^{\ell r}_{in}$,
which measures the overall similarity between the $i$-th visual features and the embeddings of normal/abnormal textual prompts. The probability that a visual token is anomalous can be calculated by
\begin{align}
    p_{i} = \frac{\exp(S_{ia})}{\exp(S_{ia}) + \exp(S_{in})}.
\end{align}
Among all $p_i$ with $i\in \{CLS, 1, 2, \cdots, N\}$, the $p_{CLS}$ can be used for anomaly classification, while probability $p_i$ from other patch $i\in \{1, 2,\cdots,N\}$ can be used for fine-grained anomaly segmentation. 

For any input image $x$, assume the image-level anomaly label is $y \in \{0, 1\}$ and the pixel-level ground truth mask is $M \in \mathbb{R}^{\tilde{H} \times \tilde{W}}$. Due to the imbalanced number of normal and anomalous samples in the auxiliary dataset, we use focal loss
\begin{align}
L_{cls} = FocalLoss(p_{CLS}, y)
\label{eq:cls_loss}
\end{align}
for anomaly classification. This loss will encourage the visual [CLS] token to attend to anomalous patterns.
For segmentation supervision, we reshape the scores $[p_1, p_2, \dots, p_N]$ into $\tilde{P} \in \mathbb{R}^{H \times W}$ as
\begin{equation} \label{Eq:seg_P}
    \tilde{P} = {\text{Reshape}}([p_1, p_2, \dots, p_N]),
\end{equation}
and resize ground truth $M$ into $\tilde{M} \in \mathbb{R}^{H \times W}$, where $H = W = \sqrt{N}$. Considering the imbalance of normal and anomalous pixels as well as the sparsity of small anomalous masks, we calculate the segmentation loss by the focal loss and L1 loss, denoted as
\begin{align}
    L_{seg} = FocalLoss(\tilde{P}, \tilde{M}) + L1(\tilde{P}, \tilde{M})
\label{eq:seg_loss}
\end{align}
Through this segmentation loss, the patch features develop strong self-attention capabilities for their anomalous characteristics, significantly improving the model's ability to localize defects.

\paragraph{Patch Alignment Loss}
Building upon the standard classification and segmentation losses, we recognize that patch features in anomalous images naturally fall into two distinct categories: normal and abnormal regions.  To improve the discriminativeness between normal and abnormal features, we formulate a patch alignment loss that operates on three types of feature pairs: normal-normal (N-N), normal-anomaly (N-A), and anomaly-anomaly (A-A). The key intuition is that N-A pairs should show lower similarity than N-N and A-A pairs. This leads to the following patch alignment loss:
\begin{align}
    L_{pal} = & \frac{1}{|\mathcal{L}||\mathcal{R}|} \sum_{\ell\in \mathcal{L}}\sum_{r\in \mathcal{R}} \max (0, \frac{1}{N_n \times N_a} \sum_{i=1}^{N_n}  \sum_{j=1}^{N_a} \langle z^{\ell r}_{ni}, z^{\ell r}_{aj} \rangle \nonumber \\
    &- \frac{1}{N_n^2 + N_a^2}(\sum_{i=1}^{N_n} \sum_{j=1}^{N_n} \langle z^{\ell r}_{ni}, z^{\ell r}_{nj} \rangle + \sum_{i=1}^{N_a} \sum_{j=1}^{N_a} \langle z^{\ell r}_{ai}, z^{\ell r}_{aj} \rangle)),
    \label{eq:patch_alignment}
\end{align}
where $N_n$ and $N_a$ denote the number of normal and abnormal patches, respectively.
The loss encourages the similarity of N-A pair $(z^{\ell r}_{ni}, z^{\ell r}_{aj})$ to be no larger than N-N and A-A pairs, thereby constraining patch features to form compact and discriminative spaces for normal and abnormal visual features.

Combining the image-level classification loss\eqref{eq:cls_loss}, pixel-level segmentation loss,\eqref{eq:seg_loss} and the patch alignment loss \eqref{eq:patch_alignment}, the final loss for optimizing AF-CLIP can be 
\begin{align}
    L = L_{cls} + \lambda_1 L_{seg} + \lambda_2 L_{pal},
\end{align}
where $\lambda_1, \lambda_2 > 0$ are the  coefficients.

\begin{table*}
\centering
\caption{Zero-shot anomaly detection performance on industrial and medical domains. The image-level performance AUROC(\%), AP(\%) and pixel-level performance AUROC(\%), PRO(\%) are reported.}
\begin{tabular}{ccccccccccccccc}
\toprule
\multirow{2}{*}{Domain}& \multirow{2}{*}{Metric} & \multirow{2}{*}{Dataset} & WinCLIP\cite{jeong2023winclip} & VAND\cite{chen2023VAND} & AnomalyCLIP\cite{zhou2024anomalyclip} & AdaCLIP\cite{cao2024adaclip} & AA-CLIP\cite{ma2025aaclipenhancingzeroshotanomaly} & AF-CLIP \\
& & & CVPR'2023 & CVPR'2023 & ICLR'2024 & ECCV'2024 & CVPR'2025 & (ours)\\
\midrule
\multirow{10}{*}{Industrial} & \multirow{5}{*}{\shortstack{image-level \\ (AUROC, AP)}} & MVTec & (91.8, 96.5) & (86.1, 93.5) & (91.5, 96.2) & (92.1, 96.7) & (90.5, 94.9) & (\textbf{92.9}, \textbf{96.8}) \\
& & Visa & (78.1, 81.2) & (78.0, 81.4) & (82.1, 85.4) & (86.3, 89.2) & (84.6, 82.2) & (\textbf{88.5}, \textbf{90.0}) \\
& & BTAD & (68.2, 70.9) & (73.5, 68.6) & (88.3, 87.3) & (92.4, 96.2) & (93.8, \textbf{97.9}) & (\textbf{94.3}, 95.2) \\
& & DAGM & (91.8, 79.5) & (94.4, 83.8) & (97.5, 92.3) & (98.3, 93.7) & (93.9, 84.5) & (\textbf{98.7}, \textbf{94.6}) \\
& & DTD & (93.2, 92.6) & (86.4, 95.0) & (93.5, 97.0) & (97.0, 99.0) & (93.3, 97.8) & (\textbf{97.9}, \textbf{99.1}) \\
\cmidrule(lr){2-9}
& \multirow{5}{*}{\shortstack{pixel-level \\ (AUROC, PRO)}} & MVTec & (85.1, 
 64.6) & (87.6, 44.0) & (91.1, 81.4) & (86.3, 20.0) & (91.9, 84.6) & (\textbf{92.3},\textbf{85.7}) \\
& & Visa & (79.6, 56.8) & (94.2, 86.8) & (95.5, 87.0) & (95.8, 60.0) & (95.5, 83.0) & (\textbf{96.2}, \textbf{88.7}) \\
& & BTAD & (72.7, 27.3) & (89.3, 68.8) & (94.2, 74.8) & (94.0, 33.8) & (94.0, 69.0) & (\textbf{94.4}, \textbf{78.3}) \\
& & DAGM & (87.6, 65.7) & (82.4, 66.2) & (95.6, 91.0) & (94.5, 53.8) & (91.6, 76.5) & (\textbf{97.1}, \textbf{93.1}) \\
& & DTD & (83.9, 57.8) & (95.3, 86.9) & (97.9, 92.3) & (98.5, 72.9) & (96.4, 85.9) & (\textbf{98.6}, \textbf{93.8}) \\
\midrule
\multirow{5}{*}{Medical} & \multirow{2}{*}{\shortstack{image-level \\ (AUROC, AP)}} & BrainMRI & (96.6, 91.5) & (89.3, 90.9) & (90.3, 92.3) & (80.6, 84.4) & (91.8, 94.1) & (\textbf{95.2}, \textbf{96.3}) \\
& & Br35H & (80.5, 82.2) & (93.1, 92.9) & (94.6, 94.7) & (86.5, 86.3) & (89.4, 91.0) & (\textbf{96.7}, \textbf{96.4}) \\
\cmidrule(lr){2-9}
& \multirow{4}{*}{\shortstack{pixel-level \\ (AUROC, PRO)}} & ISIC & (93.3, 55.1) & (89.4, 77.2) & (89.4, 78.4) & (88.0, 32.2) & (93.9, 87.0) & (\textbf{94.8},\textbf{89.6}) \\
& & ClinicDB & (51.2, 13.8) & (80.5, 60.7) & (82.9, 67.8) & (81.6, 54.0) & (86.3, 67.1) & (\textbf{87.1}, \textbf{70.0}) \\
& & ColonDB & (70.3, 32.5) & (78.4, 64.6) & (81.9, \textbf{71.3}) & (79.3, 49.1) & (81.6, 65.4) & (\textbf{83.2}, 67.9) \\
& & Kvasir	& (69.7, 24.5) & (75.0, 36.2) & (78.9, 45.6) & (77.5, 41.3) & (84.0, 50.3) & (\textbf{84.5}, \textbf{57.5}) \\
\bottomrule
\end{tabular}
\label{tab:zero-shot}
\end{table*}

\subsection{Anomaly Detection under Zero-shot and Few-shot Scenarios}
\paragraph{Zero-shot Anomaly Detection} After training the adaptor and the textual prompts, AF-CLIP can be used for zero-shot anomaly detection. Specially, $p_{CLS}$ can be used as the anomaly score for image classification, while for segmentation, we up-sample $\tilde{P}$ in \eqref{Eq:seg_P} to the same size as the image by bilinear interpolation and then smooth it with a Gaussian filter to yield final anomaly score map.

\paragraph{Few-shot Anomaly Detection} To further adapt the model to scenarios where a few normal shot samples \( \mathcal{X}_n \) are available from the target domain, we extend AF-CLIP to a few-shot method AF-CLIP+ by introducing memory banks that store aggregated patch features:
\begin{align}
    B^{\ell r}\!=\!\left\{c^{\ell r}_i|i\!\in\!\left\{1,\!2,\!\dots,\!N\right\},x\!\in\!\mathcal{X}_n\right\}, \ell\!\in\!\mathcal{L}, r\!\in\!\mathcal{R},
\end{align}
where $c_i^{\ell r}$ is aggregated with spatial information as \eqref{eq:agg}. At testing, for any patch token $i$, nearest neighbor searching is performed on its visual features $c^{\ell r}_i$ to produce the anomaly score by
\begin{align}
    A^{\ell r}_i = \min_{b \in B^{\ell r}} (1 - \langle c^{\ell r}_i, b \rangle) / 2.
\end{align}
Subsequently, we average the score over all blocks and scales, and thus obtain  $A_i = \frac{1}{|\mathcal{L}||\mathcal{R}|}\sum_{\ell, r}A^{\ell r}_i$. Then, the scores $\{A_1, A_2, \dots, A_N\}$ are reshaped into $A \in \mathbb{R}^ {H \times W}$. For few-shot anomaly detection, we combine $A$ with scores output by the zero-shot model, leading to the following two scores
\begin{align}
    S_{img} = \max_i A_i + \alpha p_{CLS},\quad
    S_{map} = A + \beta \tilde{P}
\end{align}
for image-level classification and pixel-level segmentation respectively, where $\alpha \geq 0, \beta \geq 0$ are hyper-parameters.

\section{Experiments}

\subsection{Experimental Settings} 
\paragraph{Datasets and Metrics}
Under zero-shot scenarios, we conduct experiments on 10 datasets, including 5 industrial datasets: MVTec~\cite{bergmann2019mvtec}, Visa ~\cite{zou2022visa}, BTAD~\cite{mishra2021btad}, DAGM ~\cite{wieler2007dagm}, DTD-Synthetic (DTD)~\cite{mishra2021btad}, and 6 medical datasets: BrainMRI ~\cite{salehi2021multiresolution_brainmri}, Br35H~\cite{br35h}, ISIC~\cite{codella2019isic}, CVC-ColonDB~\cite{tajbakhsh2015automated_colon}, and CVC- ClinicDB~\cite{bernal2015wm_clinic}, Kvasir~\cite{jha2020kvasir}. For few-shot anomaly detection, experiments on the widely used MVTec AD and Visa datasets are conducted.
The performance of anomaly classification is evaluated by AUROC (Area Under the Receiver Operating Characteristic Curve) and AP (Average Precision), while AUROC and AUPRO (Area Under Per-Region Overlap) are provided to evaluate the segmentation performance. The reported performance is averaged across all categories for each dataset.

\paragraph{Implementation Details} We use the frozen CLIP (VIT-L/14
@336px) and train the learnable prompts as well as the additional adaptor. All Images are resized to 518, the learnable embedding lengths $M$ are set to 12, $\lambda_1=\lambda_2=1$, and $\alpha = \beta = 0.1$. Adam is used as the optimizer with the learning rate set to 1e-4. The batch size is set to 8. Under the zero-shot setting, we train the model on the industrial dataset MVTec for 2 epochs and report the test zero-shot performance on other anomaly datasets. To report the zero-shot performance on MVTec, we train the model on Visa for 2 epochs.

\subsection{Zero-shot Anomaly Detection} 
\paragraph{Zero-shot Anomaly Detection on Industrial Datasets} Table \ref{tab:zero-shot} first shows the zero-shot performance of 5 industrial datasets. For fair comparison, methods that need optimization are trained on the same auxiliary dataset. Compared to recent state-of-the-art methods, AF-CLIP achieves superior performance. Compared to recent AdaCLIP, our method improves the image-level AUROC of 2.2\%, 1.9\%, and 0.9\% on Visa, BTAD, and DTD-Synthetic, respectively. Compared to AnomlayCLIP, AF-CLIP improves pixel-level AUROC of 1.2\%, 1.5\% and 0.7\% on MVTec, DAGM and DTD-Synthetic, respectively. The weak performance of other methods can be attributed to the insufficient adaptation of visual features. WinCLIP directly takes original pre-trained visual features and VAND just adapts simple linear projection of visual features.  AnomalyCLIP uses diagonal attention to focus on the local information of each patch but ignores the interactive information of surrounding patches and various sizes of anomalies. AdaCLIP and AA-CLIP train additional parameters for the vision encoder without explicitly optimizing for local anomaly awareness. Instead, the proposed AF-CLIP fundamentally enhances the visual features to focus on local anomalies by spatial aggregation and additional adaptor, outperforming other methods on these industrial datasets.

\paragraph{Generalization on Medical Datasets} Since the zero-shot model is trained on the public industrial datasets, to evaluate the generalization, we apply it to detect the anomalies from medical datasets. Table \ref{tab:zero-shot} also exhibits the anomaly classification performance on BrainMRI and Br35H, and anomaly localization performance on ISIC,  CVC-ColonDB, CVC-ClinicDB, and Kvasir. It can be seen that the model tuned on a general industrial dataset is also applicable to detect anomalies in the medical domain. Among all the compared methods, it is notable that AF-CLIP has achieved better image-level performance compared to other state-of-the-art zero-shot anomaly detection methods, as well as gets competitive localization results, especially on pixel-level AUROC. This can be attributed to our proposed method learns to identify general anomalies in images.

\begin{table*}[htbp]
    \centering
     \caption{Few-shot anomaly detection performance on MVTec and Visa. The average AUROC$\pm$std(\%), AP$\pm$std(\%) are reported for image-level classification while the average AUROC$\pm$std(\%) and PRO$\pm$std(\%) are measured for pixel-level segmentation.}
    \begin{tabular}{ccccccccccc}
        \toprule
        \multirow{2}{*}{Setup} & \multirow{2}{*}{Method} & \multicolumn{4}{c}{MVTec} & \multicolumn{4}{c}{Visa} \\
        \cmidrule(lr){3-6} \cmidrule(lr){7-10}
        & & I-AUROC & I-AP & P-AUROC & P-PRO & I-AUROC & I-AP & P-AUROC & P-PRO \\
        \midrule
        \multirow{6}{*}{0-shot}
        & WinCLIP\cite{jeong2023winclip} (CVPR'2023)  & 91.8$_{\pm0.0}$ & 96.5$_{\pm0.0}$ & 85.1$_{\pm0.0}$ & 64.6$_{\pm0.0}$ & 78.1$_{\pm0.0}$ & 81.2$_{\pm0.0}$ & 79.6$_{\pm0.0}$ & 56.8$_{\pm0.0}$ \\
        & AnomalyCLIP\cite{zhou2024anomalyclip} (ICLR'2024) & 91.5$_{\pm0.0}$ & 96.2$_{\pm0.0}$ & 91.1$_{\pm0.0}$ & 81.4$_{\pm0.0}$ & 82.1$_{\pm0.0}$ & 85.4$_{\pm0.0}$ & 95.5$_{\pm0.0}$ & 87.0$_{\pm0.0}$ \\
        & AdaCLIP\cite{cao2024adaclip} (CVPR'2024) & 92.1$_{\pm0.0}$ & 96.7$_{\pm0.0}$ & 86.3$_{\pm0.0}$ & 20.0$_{\pm0.0}$ & 86.3$_{\pm0.0}$ & 89.2$_{\pm0.0}$ & 95.8$_{\pm0.0}$ & 60.0$_{\pm0.0}$ \\
        & SimCLIP\cite{deng2024simclip} (ACM MM'2025) & 90.0$_{\pm0.0}$ & 95.3$_{\pm0.0}$ & 91.8$_{\pm0.0}$ & 86.8$_{\pm0.0}$ & 83.1$_{\pm0.0}$ & 86.0$_{\pm0.0}$ & 89.7$_{\pm0.0}$ & 95.6$_{\pm0.0}$ \\
        & AA-CLIP\cite{ma2025aaclipenhancingzeroshotanomaly} (CVPR'2025) & 90.5$_{\pm0.0}$ & 94.9$_{\pm0.0}$ & 91.9$_{\pm0.0}$ & 84.6$_{\pm0.0}$ & 84.6$_{\pm0.0}$ & 82.2$_{\pm0.0}$ & 95.5$_{\pm0.0}$ & 83.0$_{\pm0.0}$ \\
        & AF-CLIP(ours) & \textbf{92.9}$_{\pm0.0}$ & \textbf{96.8}$_{\pm0.0}$ & \textbf{92.3}$_{\pm0.0}$ & \textbf{85.7}$_{\pm0.0}$ & \textbf{88.5}$_{\pm0.0}$ & \textbf{90.0}$_{\pm0.0}$ & \textbf{96.2}$_{\pm0.0}$ & \textbf{88.7}$_{\pm0.0}$\\
        \midrule
        \multirow{6}{*}{1-shot} 
        & WinCLIP+\cite{jeong2023winclip} (CVPR’2023) & 93.1$_{\pm2.0}$ & 96.5$_{\pm0.9}$ & 95.2$_{\pm0.5}$ & 87.1$_{\pm1.2}$ & 83.8$_{\pm4.0}$ & 85.1$_{\pm4.0}$ & 96.4$_{\pm0.4}$ & 85.1$_{\pm2.1}$ \\
        & PromptAD\cite{li2024promptad} (CVPR’2024) & 94.6$_{\pm1.7}$ & 97.1$_{\pm1.0}$ & 95.9$_{\pm0.5}$ & 87.9$_{\pm1.0}$ & 86.9$_{\pm2.3}$ & 88.4$_{\pm2.6}$ & 96.7$_{\pm0.4}$ & 85.1$_{\pm2.5}$ \\
        & AnomalyGPT\cite{gu2024anomalygpt} (AAAI'2024) & 94.1$_{\pm1.1}$ & - & 95.3$_{\pm0.1}$ & - &87.4$_{\pm0.8}$ & - & 96.2$_{\pm0.1}$ & -\\
        & SimCLIP\cite{deng2024simclip} (ACM MM'2024) & 95.3$_{\pm0.3}$ & 97.7$_{\pm0.3}$ & 95.6$_{\pm0.2}$ & 92.4$_{\pm0.2}$ & \textbf{93.0}$_{\pm1.1}$ & \textbf{94.5}$_{\pm0.9}$ & 97.4$_{\pm0.1}$ & \textbf{92.7}$_{\pm0.2}$ \\
        & IIPAD\cite{lvone} (ICLR'2025) & 94.2$_{\pm-}$ & 97.2$_{\pm-}$ & 96.4$_{\pm-}$ & 89.8$_{\pm-}$ & 85.4$_{\pm-}$ & 87.5$_{\pm-}$ & 96.9$_{\pm-}$ & 87.3$_{\pm-}$ \\
        & AF-CLIP+(ours) & \textbf{95.5}$_{\pm1.3}$ & \textbf{98.0}$_{\pm0.5}$ & \textbf{96.8}$_{\pm0.4}$ & \textbf{92.6}$_{\pm0.8}$ & 90.5$_{\pm2.1}$ & 91.2$_{\pm1.8}$ & \textbf{98.2}$_{\pm0.1}$ & 92.5$_{\pm0.7}$ \\
        \midrule
        \multirow{6}{*}{2-shot} 
        & WinCLIP+\cite{jeong2023winclip} (CVPR'2023) & 94.4$_{\pm1.3}$ & 97.0$_{\pm0.7}$ & 96.0$_{\pm0.3}$ & 88.4$_{\pm0.9}$ & 84.6$_{\pm2.4}$ & 85.8$_{\pm2.7}$ & 96.8$_{\pm0.3}$ & 86.2$_{\pm1.4}$ \\
        & PromptAD\cite{li2024promptad} (CVPR'2024) & 95.7$_{\pm1.5}$ & 97.9$_{\pm0.7}$ & 96.2$_{\pm0.3}$ & 88.5$_{\pm0.8}$ & 88.3$_{\pm2.0}$ & 90.0$_{\pm2.1}$ & 97.1$_{\pm0.3}$ & 85.8$_{\pm2.1}$ \\
         & AnomalyGPT\cite{gu2024anomalygpt} (AAAI'2024) & 95.5$_{\pm0.8}$ & - & 95.6$_{\pm0.2}$ & - & 87.4$_{\pm0.8}$ & - & 96.2$_{\pm0.1}$ & - \\
        & SimCLIP\cite{deng2024simclip} (ACM MM'2024) & 96.0$_{\pm0.2}$ & 98.1$_{\pm0.1}$ & 96.0$_{\pm0.2}$ & 92.9$_{\pm0.1}$ & \textbf{93.7}$_{\pm0.2}$ & \textbf{94.9}$_{\pm0.2}$ & 97.7$_{\pm0.1}$ & \textbf{93.4}$_{\pm0.0}$ \\
        & IIPAD\cite{lvone} (ICLR'2025) & 95.7$_{\pm-}$ & 97.9$_{\pm-}$ & 96.7$_{\pm-}$ & 90.3$_{\pm-}$ & 86.7$_{\pm-}$ & 88.6$_{\pm-}$ & 97.2$_{\pm-}$ & 87.9$_{\pm-}$ \\
        & AF-CLIP+(ours) & \textbf{96.5}$_{\pm1.3}$ & \textbf{98.4}$_{\pm0.5}$ & \textbf{97.1}$_{\pm0.3}$ & \textbf{93.1}$_{\pm0.7}$ & 91.2$_{\pm1.4}$ & 91.9$_{\pm1.4}$ & \textbf{98.4}$_{\pm0.1}$ & 92.9$_{\pm0.4}$ \\
        \midrule
        \multirow{6}{*}{4-shot} 
        & WinCLIP+\cite{jeong2023winclip} (CVPR'2023) & 95.2$_{\pm1.3}$ & 97.3$_{\pm0.6}$ & 96.2$_{\pm0.3}$ & 89.0$_{\pm0.8}$ & 87.3$_{\pm1.8}$ & 88.8$_{\pm1.8}$ & 97.2$_{\pm0.2}$ & 87.6$_{\pm0.9}$ \\
        & PromptAD\cite{li2024promptad} (CVPR'2024) & 96.6$_{\pm0.9}$ & 98.5$_{\pm0.5}$ & 96.5$_{\pm0.2}$ & 90.5$_{\pm0.7}$ & 89.1$_{\pm1.7}$ & 90.8$_{\pm1.3}$ & 97.4$_{\pm0.3}$ & 86.2$_{\pm1.7}$ \\
        & AnomalyGPT\cite{gu2024anomalygpt} (AAAI'2024) & 95.5$_{\pm0.8}$ & - & 95.6$_{\pm0.2}$ & - & 88.6$_{\pm0.7}$ &- & 96.4$_{\pm0.1}$ & - \\
        & SimCLIP\cite{deng2024simclip} (ACM MM'2024) & 96.4$_{\pm0.2}$ & 98.0$_{\pm0.2}$ & 96.2$_{\pm0.1}$ & 93.1$_{\pm0.2}$ & \textbf{94.4}$_{\pm0.1}$ & \textbf{95.6}$_{\pm0.1}$ & 98.0$_{\pm0.2}$ & \textbf{94.1}$_{\pm0.1}$\\
        & IIPAD\cite{lvone} (ICLR'2025) & 96.1$_{\pm-}$ & 98.1$_{\pm-}$ & 97$_{\pm-}$ & 91.2$_{\pm-}$ & 88.3$_{\pm-}$ & 89.6$_{\pm-}$ & 97.4$_{\pm-}$ & 88.3$_{\pm-}$ \\
        & AF-CLIP+(ours) & \textbf{97.2}$_{\pm0.9}$ & \textbf{98.6}$_{\pm0.5}$ & \textbf{97.6}$_{\pm0.2}$ & \textbf{93.8}$_{\pm0.6}$ & 92.2$_{\pm1.3}$ & 92.5$_{\pm1.2}$ & \textbf{98.5}$_{\pm0.1}$ & 93.4$_{\pm0.5}$ \\
        \bottomrule
    \end{tabular}
    \label{tab:few-shot}
\end{table*}

\begin{figure*}[htbp]
    \centering
    \begin{minipage}[b]{0.7\textwidth}
        \centering
        \begin{subfigure}{0.6\linewidth}
            \centering
            \includegraphics[width=1.0\linewidth]{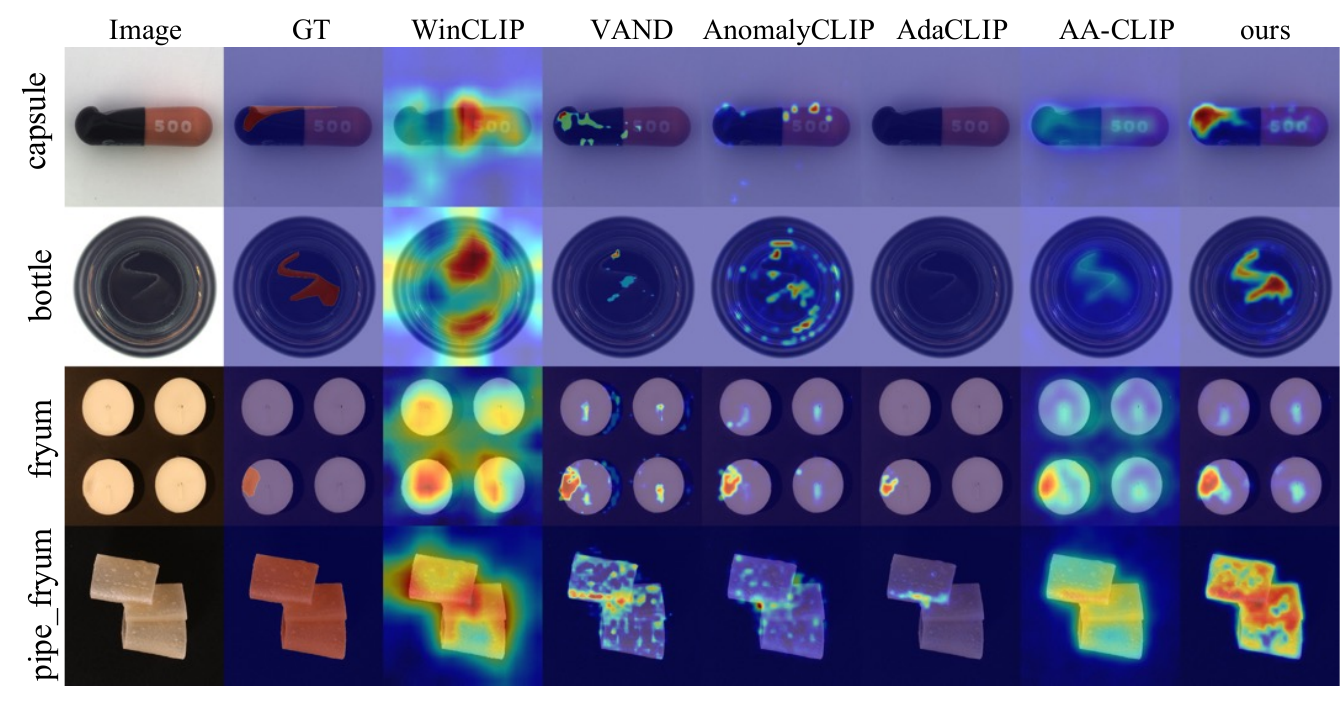}
            \caption{0-shot}
            \label{fig:0-shot}
        \end{subfigure}
        \centering
        \begin{subfigure}{0.38\linewidth}
            \centering
            \includegraphics[width=1.0\linewidth]{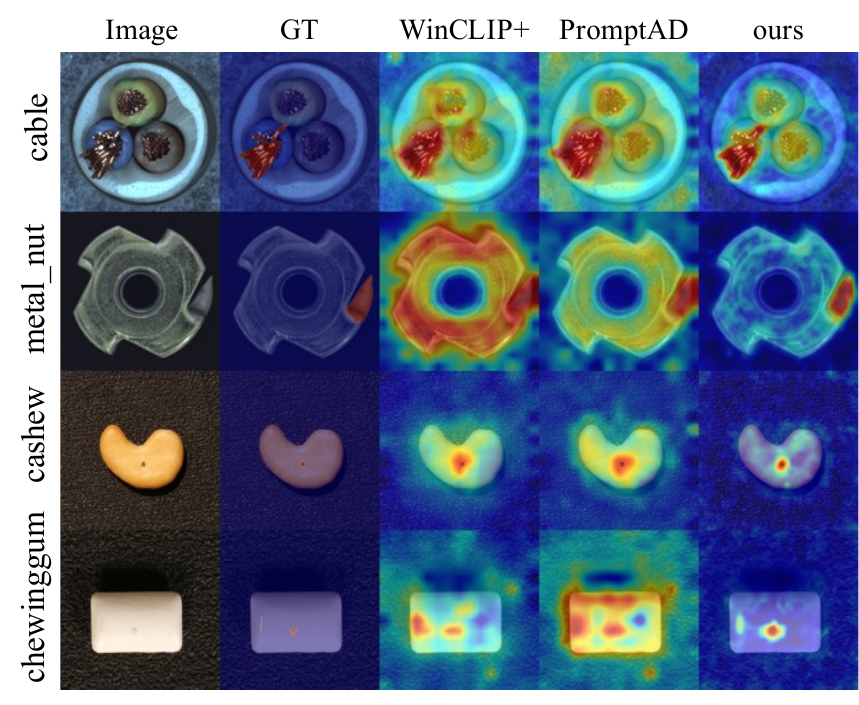}
            \caption{1-shot}
            \label{fig:1-shot}
        \end{subfigure}
        \caption{Visualization of anomaly localization results. From left to right, the first two columns show images and the ground truth (GT) of anomaly masks, other columns show the anomaly heatmaps of different methods.}
        \label{fig:result}
    \end{minipage}
    \hfill
    \begin{minipage}[b]{0.28\textwidth}
        \centering
         \captionof{table}{Comparison with many-shot methods on MVTec.}
        \resizebox{1.0\textwidth}{!}{
        \begin{tabular}{ccccccc}
        \toprule
           \multirow{2}{*}{Method}  & \multirow{2}{*}{Setting} & Image & Pixel \\
           & & AUROC & AUROC \\
           \midrule
           \multirow{4}{*}{AF-CLIP+}  & 0-shot & 92.9 & 92.3 \\
           & 1-shot & 95.5 & 96.8 \\
           & 2-shot & 96.5 & 97.1 \\
           & 4-shot & 97.2 & 97.6 \\
           \midrule
           DiffNet~\cite{rudolph2021diffNet} & 16-shot & 87.3 & -\\
           RegAD~\cite{huang2022RegAD} & 8-shot & 91.2 & 96.7 \\
           FastRecon~\cite{fang2023fastrecon} & 8-shot & 95.2 & 97.3 \\
           InCTRL~\cite{zhu2024inctrl} & 8-shot & 95.3 & - \\
           \midrule
           P-SVDD~\cite{yi2020p-svdd} & full-shot & 95.2 & 96.0 \\
           PatchCore~\cite{roth2022patchcore} & full-shot & 99.1 & 98.1 \\
           DesTeg~\cite{zhang2023destseg} & full-shot & 98.6 & 97.9 \\
           D3AD~\cite{Tebbe_2024_CVPR_D3AD} & full-shot & 97.2 & 97.4 \\
        \bottomrule
        \end{tabular}}
        \label{tab:many-shot}
    \end{minipage}
    \Description{}
\end{figure*}

\subsection{Few-shot Anomaly Detection} 

\paragraph{Few-shot Detection Results} The experimental results of different few-shot settings on the MVTec AD and VisA datasets are presented in Table \ref{tab:few-shot}, where we also report the zero-shot performance. Among all methods, AF-CLIP achieves the best zero-shot results.
For few-shot anomaly classification and segmentation, our detection model needs no further training. Compared to the no training method WinCLIP+~\cite{jeong2023winclip}, it can be obviously found that AF-CLIP+ outperforms it by a large margin.
Notably, AF-CLIP+ also demonstrates superiority over methods that leverage collected few-shot normal samples for training, such as PromptAD \cite{li2024promptad} and IIPAD \cite{lvone}. Compared with the recent SimCLIP \cite{deng2024simclip}, AF-CLIP outperforms it across all metrics on MVTec and achieves a higher pixel-level AUROC on VisA, thus exhibiting competitive few-shot performance.

\paragraph{Compared With Many-shot Methods}
As Table \ref{tab:many-shot} exhibits, compared with methods DiffNet~\cite{rudolph2021diffNet}, RegAD~\cite{huang2022RegAD}, FastRecon~\cite{fang2023fastrecon}, and InCTRL~\cite{zhu2024inctrl}, which use more shot normal samples, AF-CLIP+ outperforms them when only using 4-shot normal samples. Compared with full-shot methods P-SVDD~\cite{yi2020p-svdd}, DesTseg~\cite{zhang2023destseg}, and D3AD~\cite{Tebbe_2024_CVPR_D3AD}, the few-shot detection performance of AF-CLIP+ is also competitive in both anomaly classification and segmentation.

\begin{table*}[tbp]
    \centering
    \caption{Performance of ablation on different modules for zero-shot anomaly detection.}
    \begin{tabular}{cccccccccc}
    \toprule
        & \multirow{3}{*}{\shortstack{learnable \\ prompt }} & \multirow{3}{*}{adaptor} & \multirow{3}{*}{\shortstack{multi-level \\ features}} & \multirow{3}{*}{\shortstack{spatial \\ aggregation}} & \multicolumn{2}{c}{MVTec} & \multicolumn{2}{c}{Visa} \\
        \cmidrule(lr){6-7} \cmidrule(lr){8-9}
        & & & & & image-level & pixel-level & image-level & pixel-level \\
        & & & & & (AUROC,AP) & (AUROC, PRO) & (AUROC,AP) & (AUROC, PRO) \\
        \midrule
        base &$\times$ & $\times$ & $\times$ & $\times$ & (87.7, 94.5) & (20.0, 2.2) & (78.1, 81.6) & (26.0, 3.6) \\
        V1 &  \checkmark & $\times$ & $\times$ & $\times$ & (67.6, 85.9) & (79.6, 50.5) & (63.0, 69.3) & (89.4, 70.8) \\
        V2 & \checkmark & \checkmark  & $\times$ & $\times$ & (88.7, 94.6) & (90.2, 79.3) & (84.4, 86.8) & (94.7, 86.3) \\
        V3 & \checkmark & \checkmark & \checkmark & $\times$ & (91.4, 96.1) & (91.9, 83.6) & (86.5, 88.4) & (96.0, 88.4) \\
        V4 & \checkmark & \checkmark & \checkmark & \checkmark & (\textbf{92.9}, \textbf{96.8}) & (\textbf{92.3}, \textbf{85.7}) & (\textbf{88.5}, \textbf{89.9}) & (\textbf{96.2}, \textbf{88.7}) \\
    \bottomrule
    \end{tabular}
    \label{tab:ablation}
\end{table*}

\begin{table*}[tbp] 
    \centering
    \begin{minipage}[t]{0.6\textwidth}
        \centering
        \captionof{table}{Different fine-tuning ways of vision encoder. Image-level AUROC (I) and Pixel-level AUROC (P) are reported.}
         \resizebox{0.95\textwidth}{!}{
        \begin{tabular}{ccccccccc}
        \toprule
        \multirow{2}{*}{\shortstack{Fine-\\tune}} & \multicolumn{2}{c}{MVTec}  & \multicolumn{2}{c}{Visa} & \multirow{2}{*}{\shortstack{Learnable\\ Param}} & \multirow{2}{*}{\shortstack{Training \\ Memory}} &  \multirow{2}{*}{\shortstack{Infer \\ Time}} &  \multirow{2}{*}{GFLOPs} \\
        \cmidrule(lr){2-3} \cmidrule(lr){4-5}
        & I & P & I & P & \\
        \midrule
        Prompt    & 65.3 & 83.9 & 67.8 & 94.0 & 21504 & 9128MiB & 0.139s & 287.778\\
        Prefix   &  72.4 & 87.5 & 75.9 & 94.2 & 70656 & 22910MiB & 0.285s & 836.997\\
        LoRA & 90.5 & 91.5 & 85.7 & 96.1 & 2106368 & 8134MiB & \textbf{0.138s} & \textbf{285.362}\\
        Adaptor & \textbf{92.9} & \textbf{92.3} & \textbf{88.5} & \textbf{96.2} & 2109440 & \textbf{3862MiB} & 0.142s & 293.998 \\
        \bottomrule
        \end{tabular}}
        \label{tab:fine-tune}
    \end{minipage}
    \quad
    \begin{minipage}[t]{0.36\textwidth}
        \captionof{table}{Ablation of patch alignment loss. Image-level AUROC (I) and Pixel-level AUROC (P) are reported.}
        \resizebox{1.0\textwidth}{!}{
        \begin{tabular}{cccccc}
        \toprule
        \multirow{2}{*}{Loss} & \multicolumn{2}{c}{MVTec} & \multicolumn{2}{c}{Visa} \\
        \cmidrule(lr){2-3} \cmidrule(lr){4-5}
        & I & P & I & P \\
        \midrule
        $L_0 = L_{cls} + \lambda_1 L_{seg}$ & 92.4 & 91.8 &  87.9 & 95.8 \\
        $L_1=L_0 + \lambda_2L_{pal}$  & \textbf{92.9} & \textbf{92.3} & \textbf{88.5} & \textbf{96.2} \\
        \bottomrule
        \end{tabular}}
        \label{tab:loss}
    \end{minipage}
\end{table*}

\subsection{Visualization} 
Further,  Figure \ref{fig:result} illustrates the anomaly segmentation results under zero-shot and 1-shot.  As Figure\ref{fig:result}(a) shows, our proposed method can accurately localize anomalous areas in the inspected objects, while other methods either fail to identify anomalies or produce noisy segmentation maps. For 1-shot anomaly detection, our method also has a better anomaly segmentation capability.

\subsection{Ablation Study}

\paragraph{Component Analysis}Table~\ref{tab:ablation} evaluates the contributions of key architectural components in our method: (1) learnable textual prompts, (2) multi-level visual feature extraction, (3) spatial information aggregation, and (4) the self-attention adapter module.
The baseline (Base) uses fixed textual prompts ("a photo of an object without/with defect") and visual features solely extracted from the final encoder block. While this simple adaptation achieves reasonable image-level classification performance (87.7\% AUROC on MVTec), its pixel-level performance is limited (20.0\% AUROC), indicating poor localization.
V1 introduces learnable textual prompts, which yields an interesting trade-off: while image-level AUROC decreases significantly (87.7\% $\rightarrow$ 67.6\% on MVTec), pixel-level AUROC improves (20.0\% $\rightarrow$ 79.6\% ). This suggests that learned prompts better capture fine-grained patch characteristics crucial for segmentation
V2 adds a trainable adapter to enable more effective feature adaptation. Compared to V1, V2 achieves substantial improvements on MVTec in both image-level AUROC (67.6\% → 88.7\%) and pixel-level AUROC (79.6\% → 90.2\%) performance, demonstrating the adapter's ability to extract anomaly-sensitive visual representations.
V3 incorporates multi-level features from different encoder blocks, providing access to richer visual information across multiple scales. This enhancement leads to consistent performance gains across all metrics.
Our full model (V4) further enhances performance through spatial feature aggregation. The consistent improvements over V3 (e.g., MVTec image-level AUROC: 91.4\% → 92.9\%) validate the effectiveness of spatial information aggregation in detecting anomalies.

\paragraph{Visual Fine-tune Ways}


As shown in Table~\ref{tab:fine-tune}, we investigate the effectiveness of different fine-tuning strategies for adapting CLIP to zero-shot anomaly detection. Specifically, we fine-tune the vision encoder using four techniques: prompt tuning~\cite{lester2021prompt}, prefix-tuning~\cite{li2021prefix}, LoRA~\cite{hu2022lora}, and the Adaptor used in AF-CLIP, while keeping all other modules consistent with AF-CLIP.
The learnable prompts and prefixes are set to a length of 12, with the prefix layer depth fixed at 4. For LoRA, trainable parameters are inserted into the final layers of the four blocks in the vision encoder, with a rank of 64.
Results indicate that among all fine-tuning techniques, the adaptor in AF-CLIP yields the best performance. Additionally, Table~\ref{tab:fine-tune} compares the number of trainable parameters, training memory consumption (with a batch size of 1), inference time, and computational complexity. Notably, although prompt tuning and prefix tuning require fewer trainable parameters, they consume more training memory due to deeper gradient backpropagation. Similarly, LoRA demands more memory than the adaptor despite having a comparable number of trainable parameters. Our adaptor in AF-CLIP introduces minimal overhead (in terms of inference time and computational complexity) compared to LoRA, while achieving superior performance and training memory efficiency.

\paragraph{Loss Function Analysis}
Table~\ref{tab:loss} demonstrates the effectiveness of our proposed patch alignment loss ($L_{pal}$). The experimental results show that adding $L_{pal}$ simultaneously improves both pixel-level segmentation accuracy and image-level classification performance. We attribute this dual improvement to (1) the patch alignment mechanism's ability to optimize the patch feature space and (2) enhance the learning of generic textual state features by the visual-textual alignment. These combined effects lead to more robust anomaly detection at both image and pixel levels.

\section{Conclusion}

In this paper, we propose a new method for adapting CLIP to zero-/few-shot anomaly detection. To enhance the visual features to focus on local anomalies, we introduce a lightweight adaptor that refines the visual representations after incorporating spatial information aggregation to capture neighborhood context, thereby improving anomaly identification accuracy. 
Additionally, we leverage prompt tuning to learn generalized textual features of normal and abnormal states, which are then aligned with visual features to identify anomalies. Through hybrid optimization on an auxiliary dataset, visual features are directed to focus on anomalies, enabling zero-shot anomaly detection for target inspected objects.
For few-shot scenarios, we introduce memory banks to store aggregated features from few normal samples. Extensive experiments on industrial and medical datasets demonstrate the superior performance and strong generalization capability of our proposed method.

\begin{acks}
This work is supported by the National Natural Science Foundation of China (No. 62276280), Guangzhou Science and Technology Planning Project (No. 2024A04J9967). 
\end{acks}

\bibliographystyle{ACM-Reference-Format}
\bibliography{camera_ready}

\end{document}